\pdfoutput=1
\documentclass[11pt]{article}

\usepackage[preprint]{acl}

\usepackage{times}
\usepackage{latexsym}

\usepackage[T1]{fontenc}
\usepackage[utf8]{inputenc}

\usepackage{microtype}

\usepackage{inconsolata}

\usepackage{graphicx}

\usepackage{multicol}
\usepackage{multirow} 
\usepackage{amsmath}
\usepackage{booktabs}
\usepackage{bm}  
\usepackage{amssymb}
\usepackage{ragged2e}
\usepackage{subcaption}

\newcommand{\squishlisttwo}{
 \begin{list}{\scalebox{0.6}{$\bullet$}} % Scale the bullet to 50% of its original size
  { \setlength{\itemsep}{1pt}
     \setlength{\parsep}{0pt}
    \setlength{\topsep}{0pt}
    \setlength{\partopsep}{0pt}
    \setlength{\leftmargin}{1em}
    \setlength{\labelwidth}{1.5em}
    \setlength{\labelsep}{0.5em} } }

\usepackage{array}
\title{\texttt{TRACE}: \underline{TR}ansformer-based \underline{A}ttribution using \underline{C}ontrastive \underline{E}mbeddings in LLMs}

\author{Cheng Wang$^{\textbf{1}}$, Xinyang Lu$^{\textbf{1}\textbf{2}}$, See-Kiong Ng$^{\textbf{1}\textbf{2}}$, Bryan Kian Hsiang Low$^{\textbf{1}}$ \\
$^{1}$Department of Computer Science, National University of Singapore  \\
$^{2}$Institute of Data Science, National University of Singapore \\
\texttt{\{wangcheng, xinyang.lu\}@u.nus.edu} \\
\texttt{seekiong@nus.edu.sg, lowkh@comp.nus.edu.sg} \\
}

\begin{document}
\maketitle
\begin{abstract}
The rapid evolution of \emph{large language models} (LLMs) represents a substantial leap forward in natural language understanding and generation. However, alongside these advancements come significant challenges related to the accountability and transparency of LLM responses. Reliable source attribution is essential to adhering to stringent legal and regulatory standards, including those set forth by the General Data Protection Regulation. Despite the well-established methods in source attribution within the computer vision domain, the application of robust attribution frameworks to natural language processing remains underexplored. To bridge this gap, we propose a novel and versatile \underline{TR}ansformer-based \underline{A}ttribution framework using \underline{C}ontrastive \underline{E}mbeddings called \texttt{TRACE} that, in particular, exploits contrastive learning for source attribution. We perform an extensive empirical evaluation to demonstrate the performance and efficiency of \texttt{TRACE} in various settings and show that \texttt{TRACE} significantly improves the ability to attribute sources accurately, making it a valuable tool for enhancing the reliability and trustworthiness of LLMs.
\end{abstract}

\section{Introduction}
\label{sec:intro}
The recent era has seen a significant rise in the prevalence of \emph{large language models} (LLMs)~\citep{ouyang2022training,touvron2023llama} which have demonstrated an array of remarkable capabilities. However, studies ~\citep{huang2023survey,liu2024trustworthy,wang2023decodingtrust} have highlighted a critical concern on the accountability of LLMs. 
Considering the widespread usage and such a concern, it has brought to the forefront a critical need for source attribution that involves identifying the specific training data that contributes to generating part or all of an LLM's response, which is crucial for legal and regulatory compliance and enhances the reliability of LLMs.
Various regulations mandate transparency and accountability in data usage, especially regarding intellectual property and privacy. For instance, the General Data Protection Regulation (GDPR) in the European Union requires that individuals have the right to be informed when their personal data is used. Proper source attribution ensures compliance with such legal frameworks, mitigating the risk of legal disputes and penalties.

A related topic would be that of \emph{membership inference} (MI)~\citep{mireshghallah2022quantifying} whose task is to determine whether a given piece of data was used during the training of a machine learning model. While MI and source attribution share some similarities, they differ significantly in their granularity: MI typically only involves a binary classification task and does not require identifying a specific data provider. In contrast, source attribution requires to identify one or more data providers.

Though there are some studies on source attribution~\citep{marra2018gans, yu2022artificial}, a majority of them are situated within the computer vision domain. Techniques developed for computer vision tasks cannot be directly applied to LLMs due to the fundamental differences in the data and model architectures. To the best of our knowledge, effective source attribution for LLMs still remains an open and underexplored problem.

While numerous properties are important to a source attribution framework, we identify \textbf{accuracy}, \textbf{scalability}, \textbf{interpretability}, and \textbf{robustness} as the most crucial components. These four attributes are fundamental to ensuring the effectiveness and applicability of the framework across various contexts.
Accuracy is essential to guaranteeing that the framework consistently produces reliable results. Scalability ensures that the framework can handle increasing volumes of data and complexity without a significant performance degradation, making it suitable for large-scale applications. Interpretability is equally critical as it enables stakeholders to understand and trust the attribution outcomes, hence fostering transparency and facilitating informed decision making.
Robustness is vital to ensure that the framework remains reliable and effective even in the face of adversarial distortions.

This paper presents a novel \underline{TR}ansformer-based  \underline{A}ttribution framework using \underline{C}ontrastive \underline{E}mbeddings (\texttt{TRACE}) to achieve source attribution while satisfying the above three important properties. By detailing our methodology and presenting empirical results, we seek to demonstrate the accuracy, scalability, interpretability, and robustness of \texttt{TRACE}.

\begin{figure*}[t]
  \centering
  \includegraphics[width=\linewidth]{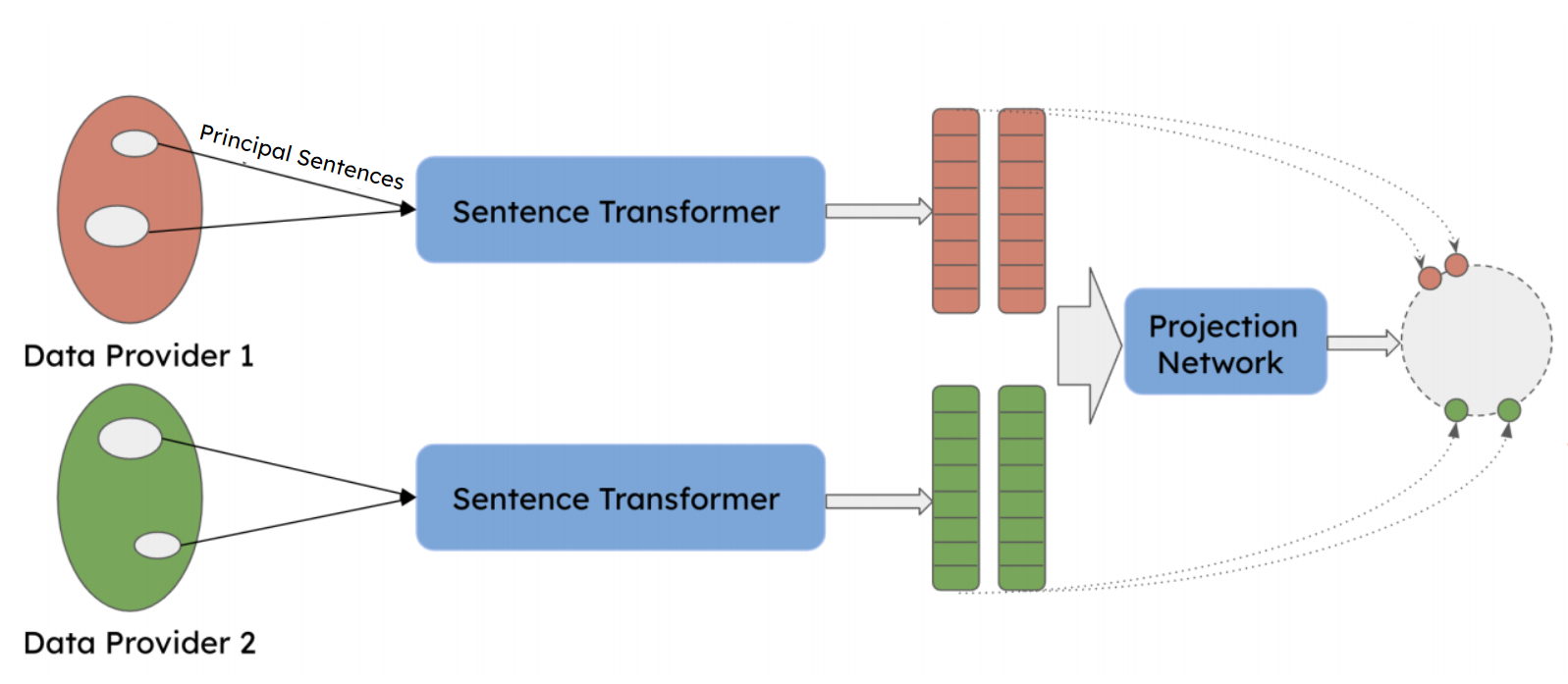}
  \caption{Illustration of \texttt{TRACE} framework.}
  \label{fig:trace_framework}
\end{figure*}

Our contributions can be summarized as follows:
\begin{itemize}
   \item We propose the novel \texttt{TRACE} framework based on contrastive learning, which is designed to achieve effective source attribution. \texttt{TRACE} differs from traditional contrastive learning by using source information as the label. Fig.~\ref{fig:trace_framework} illustrates the \texttt{TRACE} framework.

   \item We have performed an extensive empirical evaluation of \texttt{TRACE} to demonstrate its accuracy, scalability, interpretability, and robustness.
\end{itemize}

\section{Preliminaries}
\label{sec:Preliminaries}
\paragraph{Contrastive Learning and \texttt{NT-Xent Loss}.}
Contrastive learning is a conventional technique commonly used in representation learning~\citep{arora2019theoretical,inproceedings}. Its underlying idea is that similar objects should exhibit a closer distance in the embedding space while dissimilar objects should repel each other. This technique has been widely employed in computer vision tasks due to its convenient implementation to augment image input to form a self-supervised problem. Models using contrastive learning
have achieved state-of-the-art 
performances~\citep{Cui_2021_ICCV,tian2020makes}. Apart from the attention it receives in computer vision, new approaches using contrastive learning in natural language processing~\citep{meng2021cocolm, wu2020clear} have also started gaining attention and showcasing great capabilities.

Our \texttt{TRACE} framework assigns the same label to all the data from the same source, hence naturally forming a \emph{supervised contrastive learning} problem. 
In particular, 
\texttt{TRACE} utilizes \texttt{NT-Xent Loss}~\citep{NIPS2016_6b180037} for supervised contrastive learning:
\[
\mathcal L = \sum_{i \in I} \frac{-1}{|P_i|} \sum_{p \in P_i} \log\left(\frac{\exp \left( \mathbf{z}_i \cdot \mathbf{z}_p / \tau \right)}{\sum_{a \in A_i} \exp \left( \mathbf{z}_i \cdot \mathbf{z}_a / \tau \right)} \right)
\]
where the set $I$ ($P_i\subset I$) contains indices of the sentences in the given batch (sharing the same label as sentence $i$, but does not include $i$), 
$A_i = I \setminus \{i\}$, $\mathbf{z}_i$ denotes the embedding of sentence $i$, and $\tau\in\mathbb{R}^+$ is a temperature parameter. Minimizing $\mathcal L$ would maximize the similarity between embeddings (of sentences) from the same source while minimizing the similarity between embeddings from different sources.

\paragraph{Sentence Encoder.}
Similar to the concepts of \texttt{Word2Vec}~\citep{mikolov2013efficient} and \texttt{GloVe}~\citep{Brochier_2019} which produce meaningful vector representations of words, such techniques can be applied to larger text units such as sentences. A straightforward way is to take the average of word embeddings within a sentence, but this often results in embeddings that lack semantic depth. Several models have been developed to address this issue, including \texttt{InferSent}~\citep{conneau2018supervised}, \texttt{Universal Sentence Encoder}~\citep{cer2018universal}, and \texttt{Sentence-BERT} (\texttt{SBERT})~\citep{reimers2019sentencebert}. Given its superior performance and efficiency, \texttt{SBERT} is chosen to generate sentence embeddings in \texttt{TRACE}. \texttt{SBERT} leverages a pre-trained BERT network and utilizes Siamese and triplet network structures to produce semantically meaningful sentence embeddings.

\section{\texttt{TRACE} Framework}
\label{sec:method}

\subsection{Source-Specific Semantic Distillation}
\label{sec:method:key}

Projecting every piece of data from each provider into the embedding space is desirable but would incur considerable computational costs. Moreover, it is prudent to recognize that not all information carries equal importance: For example, sentences that occur less frequently typically tend to be more representative of the document. So, we propose to extract principal sentences from each source by leveraging the \emph{Term Frequency-Inverse Document Frequency} (TF-IDF) which is effective for identifying significant sentences within documents. It is generally recommended to select $10$-$20\%$ of the sentences, thereby striking a balance between complexity and performance; these sentences are subsequently defined as \emph{principal sentences}. The length of these sentences is specified by a parameter called \texttt{WINDOW\_SIZE}. Section~\ref{sec:ablation} presents an ablation study examining the effect of different \texttt{WINDOW\_SIZEs} on accuracy.

\texttt{SBERT}~\citep{reimers2019sentencebert} has proven effective in deriving high-quality sentence representations. However, to enhance its suitability for \texttt{TRACE}, we propose several modifications inspired by the work of \texttt{SimCLR}~\citep{chen2020simple}. A key finding from \texttt{SimCLR} is that adding a non-linear projection head significantly improves the representation quality. Following this insight, we incorporate a projection network at the end of the traditional \texttt{SBERT} architecture. This projection network is trained together with the base \texttt{SBERT} model, thus encouraging the learned representations to be more discriminative in the embedding space. 

\subsection{Supervised Contrastive Embedding Training for Source-Coherent Clustering}
\label{sec:method:contrastive}

Unlike the other contrastive learning frameworks in computer vision whose tasks are typically defined to be auto-regressive due to the availability of various data augmentation techniques to generate positive samples, \texttt{TRACE} aims to achieve source-coherent clustering. In our case, we already possess the label of each sentence indicating its source. So, we can frame our task as a \emph{supervised contrastive learning} problem. The supervision is derived from the label information which corresponds to the source. Contrastive learning aligns with our objective to form clusters based on these various sources.

\texttt{SimCLR} has demonstrated that \texttt{NT-Xent Loss} outperforms other contrastive loss functions such as logistic loss~\citep{mikolov2013efficient} and margin loss~\citep{Schroff_2015}. So, we employ \texttt{NT-Xent Loss} as the loss function for \texttt{TRACE}.

\subsection{Proximity-based Inference}
\label{sec:method:inference}

\begin{figure*}[t]
  \centering
  \includegraphics[width=\linewidth]{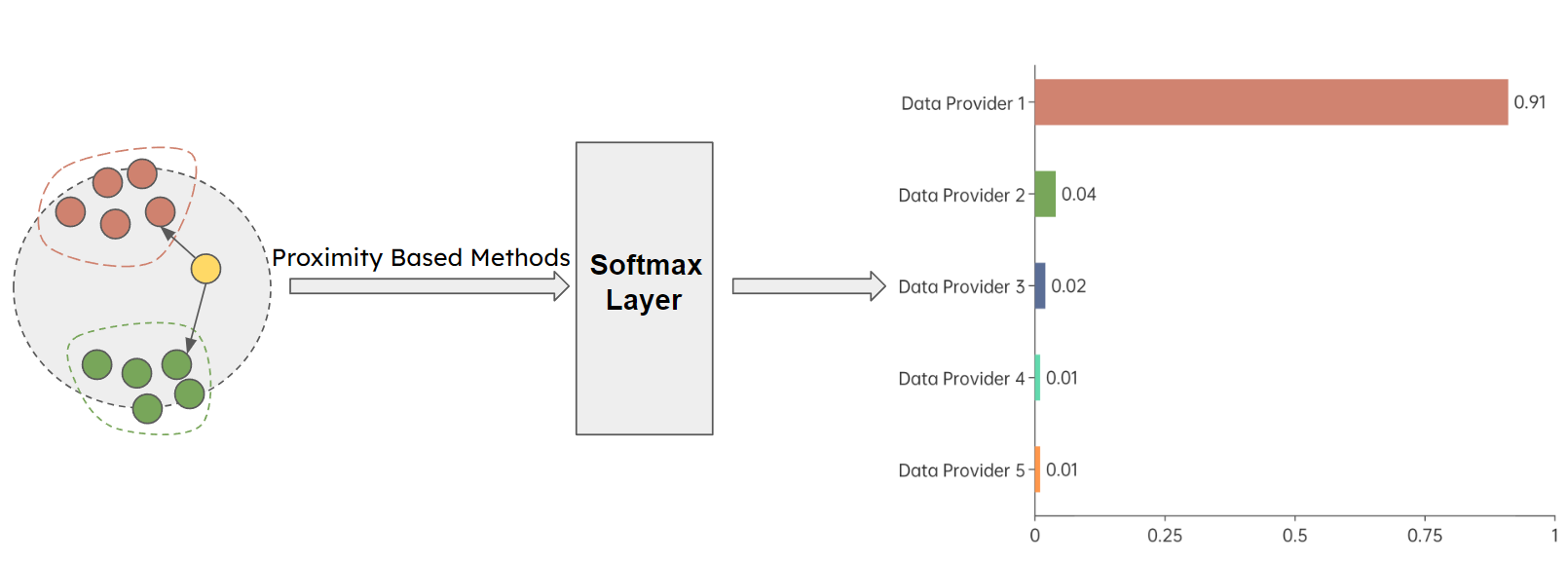}
  \caption{Illustration of the attribution step in \texttt{TRACE} framework.}
  \label{fig:trace_inference}
\end{figure*}

Once the training phase is completed, we transition to the inference stage where each data source is represented by its own set of contrastive embeddings. At this stage, when a language model generates a response, we employ the \emph{$k$-Nearest Neighbor} ($k$NN) algorithm to assign the response to the closest data source in the embedding space, as demonstrated in Fig.~\ref{fig:trace_inference}. This ensures accurate source attribution by matching the generated response with its most similar source representation.

However, responses generated by language models may not always be exclusively influenced by a single data source: there could be instances where information from multiple sources contributes to the generated text. To consider this possibility, we introduce the concept of \emph{multi-source attribution}. Multi-source attribution acknowledges and accounts for the potential influence of multiple data sources on the generated response.

We have developed three different implementations for single-source attribution and multi-source attribution, which allow users to select the most appropriate inference method based on time constraints and the number of sources. Section~\ref{sec:exp} provides a comparison of these methods.

\paragraph{Hard $k$NN (Single-Source Attribution).}
\emph{Hard $k$NN} follows the traditional $k$NN algorithm closely. Here, the attribution is determined by considering the $k$ embeddings that are closest in distance to the query. The source that appears most frequently among these $k$ neighbors is assigned as the source of the query.

\paragraph{Soft $k$NN (Multi-Source Attribution).}
To differentiate from traditional $k$NN where each query is assigned to a single source, we introduce \emph{soft $k$NN}. Here, $k$ represents the number of data sources rather than the number of closest neighbors. We rank the distances from the query to all other embeddings and select them in ascending order of distance until $k$ distinct sources are covered.

\paragraph{Nearest Centroid (Single-Source Attribution).}
To reduce inference time, we utilize the nearest centroid method. The centroid of each cluster is determined by aggregating the normalized embeddings within that cluster (i.e., corresponding to the sentences with the same label/source), as elaborated in Appendix~\ref{app:proof}.

We then apply $k$NN using these centroids. This method significantly reduces inference time as it scales with the number of data providers rather than the volume of data from each source. We will demonstrate in the next section that this method maintains an impressively high accuracy.

\section{Experiments}
\label{sec:exp}

\subsection{Experimental Setup}
\label{sec:exp:set_up}

\paragraph{Data.}

We perform an extensive empirical evaluation of \texttt{TRACE} using three datasets: \texttt{booksum}~\citep{kryscinski2021booksum}, \texttt{dbpedia\_14}~\citep{NIPS2015_250cf8b5}, and \texttt{cc\_news}~\citep{Hamborg2017}; a summary of these datasets can be found in Table~\ref{tab:dataset_stats} in the appendix. In the \texttt{booksum} dataset, we treat different books as distinct data providers and vary the number of data providers from $10$, $25$, $50$, to $100$ to demonstrate \texttt{TRACE}'s scalability to a large number of data providers. Similarly, each class in \texttt{dbpedia\_14} or each domain in \texttt{cc\_news} is considered a separate data provider. In this section, we primarily present the experimental results on the \texttt{booksum} dataset with $25$ data providers. Section~\ref{sec:additional} provides additional results.

\paragraph{Model.}
Focusing primarily on the \texttt{booksum} dataset, we evaluate the performance of \texttt{TRACE} using three different LLMs of varying sizes: \texttt{t5-small-booksum}~\citep{2020t5}, \texttt{GPT-2}~\citep{radford2019language}, and \texttt{Llama-2}~\citep{touvron2023llama}. The \texttt{t5-small-booksum} model is readily available on Hugging Face,\footnote{\href{https://huggingface.co/cnicu/t5-small-booksum}{https://huggingface.co/cnicu/t5-small-booksum}.} while \texttt{GPT-2} and \texttt{Llama-2} have been fine-tuned on a subset of the \texttt{booksum} dataset. This setup allows us to assess the performance of \texttt{TRACE} across LLMs of different scales. In App.~\ref{app:train}, we provide more details about our experiments.

\subsection{Visualization of \texttt{TRACE}'s Embedding Space}
\begin{figure*}[t]
  \includegraphics[width=0.43\linewidth]{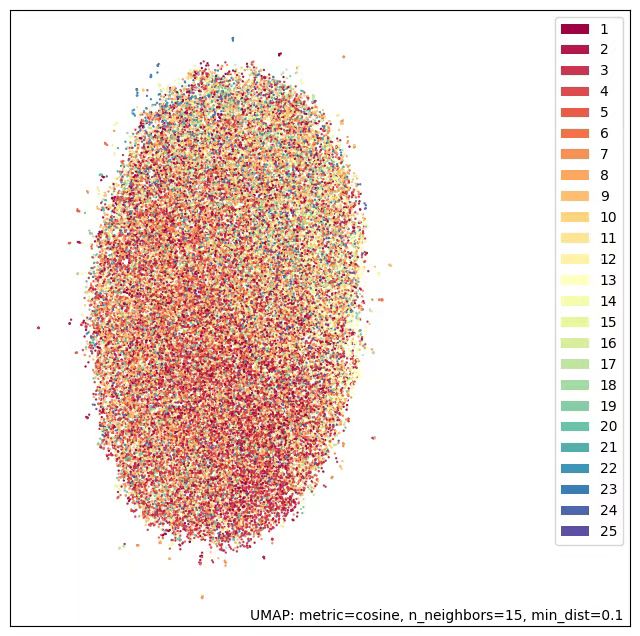} \hfill
  \includegraphics[width=0.43\linewidth]{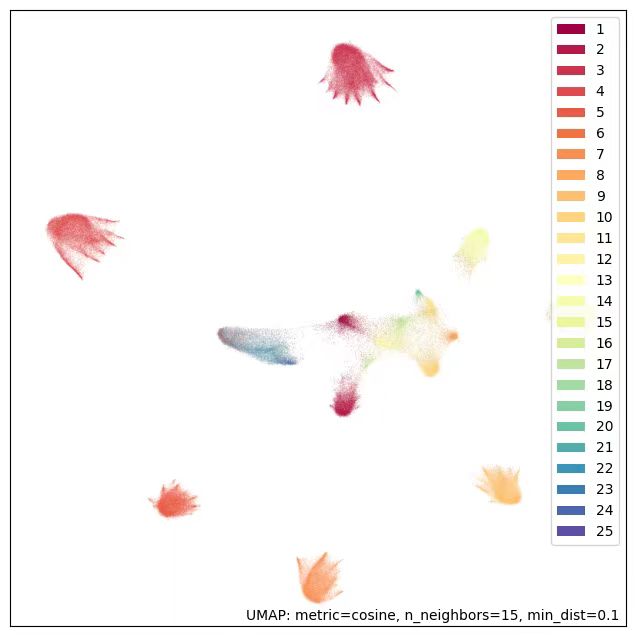}
  \caption{Visualization (using UMAP) of the embedding space before (left) and after (right) contrastive learning.}
  \label{fig:umap_visualization}\vspace{-2mm}
\end{figure*}

After training for $150$ epochs on \texttt{booksum}, a visualization tool such as \texttt{UMAP}~\citep{mcinnes2020umap} can be used to view the distribution of principal sentences. Fig.~\ref{fig:umap_visualization} shows that after the contrastive learning step, the desired outcome has been achieved, i.e., data coming from the same source form clear and distinct clusters. This validates that our contrastive learning successfully groups different data providers. Supposing the responses from an LLM are projected into the embedding space without incorporating the contrastive learning step, the resulting neighborhood exhibits chaos and it is challenging to derive robust information. This further demonstrates the importance of the contrastive learning step. 

\subsection{Accuracy}
\label{sec:exp:accuracy}
Evaluating the accuracy of source attribution is particularly challenging due to the inherent difficulty in obtaining ground-truth test datasets. Even with a dataset, a language model, and specific inputs, pinpointing the exact parts of the training data that influence a particular response remains complex. Here are the key reasons:

\begin{enumerate}
    \item \textbf{Lack of Explicit Traceability.} Language models like LLMs generate responses based on patterns learned from vast amounts of data. However, these models do not provide explicit traceability back to the specific training data. This means we cannot directly observe which parts of the training data contribute to a given response.
    
    \item \textbf{Intermixed Training Data.} The training data for LLMs is often a massive, intertwined collection of texts from various sources. Disentangling these sources to identify the precise contribution of each segment to the final response is nearly impossible due to the sheer volume and complexity.
    
    \item \textbf{Influence of Pre-training Data.} It is also likely that the model generates responses based on data encountered during the pre-training stage, which comprises a vast and diverse corpus. This pre-training data is often not fully documented or accessible, making it difficult to determine its influence on specific responses during fine-tuning or evaluation.
\end{enumerate}

Due to these challenges, obtaining ground-truth test datasets that accurately reflect the contribution of specific training data to the responses of LLMs is exceedingly difficult. To address this issue, our approach involves using training data where the source is known. We then use this known source as the ground-truth label and evaluate whether \texttt{TRACE} can correctly determine the source. This allows us to approximate the evaluation of source attribution by leveraging the known origins of the specific training data.

\paragraph{Single-Source Attribution Accuracy.}

In this case, accuracy is simply defined as the number of correct source attributions divided by the total number of attributions evaluated, the latter of which is $250$ in our experimental setup. 

\paragraph{Multi-Source Attribution Accuracy.}
In certain settings, providing multiple sources and allowing the user to determine the justification of the attribution is acceptable. For a successful \emph{soft $k$NN} attribution in such cases, the ground-truth source must appear among the \emph{top-$k$} sources returned by \texttt{TRACE}. Using the same setup as that of single-source attribution, we have evaluated \texttt{TRACE} on $250$ instances.
Table~\ref{table:booksum} below shows the results:

\begin{table}[h]
    \centering
    \resizebox{1\linewidth}{!}{
        \begin{tabular}{c|ccc|cc|c}
            \toprule
            \multirow{2}{*}{Model} & \multicolumn{3}{c|}{Soft $k$NN} & \multicolumn{2}{c|}{Hard $k$NN} & \multirow{2}{*}{Nearest Centroid}  \\
             & acc.  & top-3 acc. & top-5 acc.  & $k=10$ & $k=20$ &  \\
            \midrule
            t5 & 84.4\% & 95.3\% & 97.3\% & \multicolumn{2}{c|}{84.4\%} & 84.4\% \\ 
            \midrule
            GPT-2 & 81.3\% & 92.3\% & 94.0\% & \multicolumn{2}{c|}{81.3\%} & 81.3\% \\ 
            \midrule
            Llama-2 & 86.2\% & 96.1\% & 97.2\% & \multicolumn{2}{c|}{86.2\%} & 86.2\% \\ 
            \bottomrule 
        \end{tabular}
    }
    \caption{Source attribution accuracy for $25$ data providers on \texttt{booksum} dataset using \texttt{TRACE}.}
    \label{table:booksum}
\end{table}

It can be observed that the accuracy for models of different sizes remains consistently high and significantly surpasses the random guess' accuracy of $4\%$. Another notable observation from the results is that varying the values of $k$ in the hard $k$NN approach has minimal impact on accuracy and yields results identical to that of the nearest centroid method, which we attribute to the highly compact nature of the embeddings learned under the \texttt{TRACE} framework. When a query is projected into the embedding space, it becomes closely associated with its nearest neighbors regardless of the specific value of \( k \). This compactness suggests that the centroid of each cluster serves as an excellent representative of the entire cluster. Consequently, relying solely on these centroids can significantly reduce inference time. Even with $100$ data providers as demonstrated in next subsection, the inference process remains almost instantaneous.

\begin{table*}[t]
\centering
\resizebox{0.8\linewidth}{!}{
\begin{tabular}{c|ccc|ccc|ccc}
\toprule
\multirow{2}{*}{n\_books} & \multicolumn{3}{c|}{t5} & \multicolumn{3}{c|}{GPT2}& \multicolumn{3}{c}{Llama-2}\\
   & acc.   & top-$3$ acc. & top-$5$ acc. & acc.   & top-$3$ acc. & top-$5$ acc. & acc.   & top-$3$ acc. & top-$5$ acc.\\
\midrule

10  & 87.5\%  & 98.3\% & 99.4\% & 85.3\% & 96.8\% & 98.7\% & 88.2\% & 99.2\% & 99.5\%\\
25  & 84.4\%  & 95.3\% & 97.3\% & 81.3\% & 92.3\% & 94.0\% & 86.2\% & 96.1\% & 97.2\%\\ 
50  & 73.1\%  & 82.0\% & 84.0\% & 72.9\% & 82.9\% & 84.1\% & 70.3\% & 79.8\% & 82.2\%\\ 
100 & 45.4\%  & 74.8\% & 78.8\% & 49.0\% & 73.2\% & 77.7\% & 46.7\% & 76.8\% & 80.2\%\\ 
\bottomrule 
\end{tabular}
}

\caption{Source attribution accuracy for different no.~of data providers on \texttt{booksum} dataset using \texttt{TRACE}. 
}
\label{table: scalability}
\end{table*}

\begin{figure*}[h]
    \centering
    \begin{subfigure}[b]{0.44\linewidth}
        \centering
        \includegraphics[width=\linewidth]{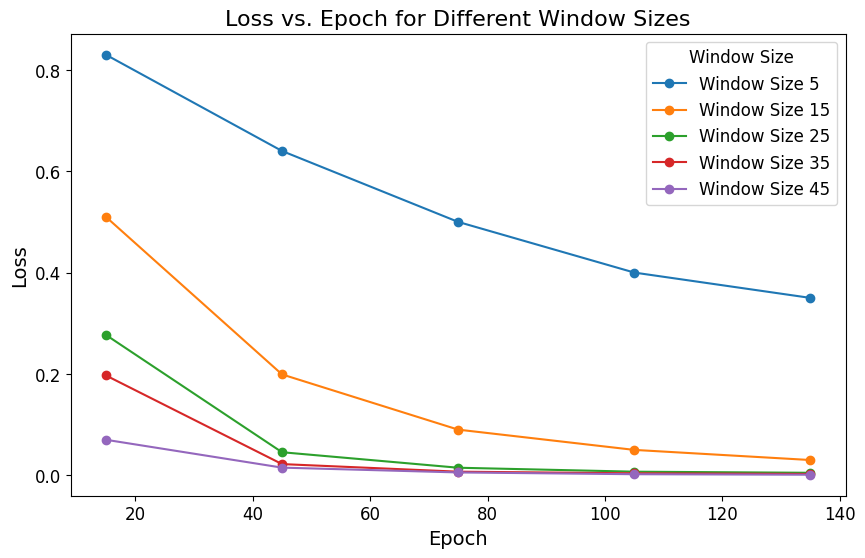}
    \end{subfigure}
    \hfill
    \begin{subfigure}[b]{0.44\linewidth}
        \centering
        \includegraphics[width=\linewidth]{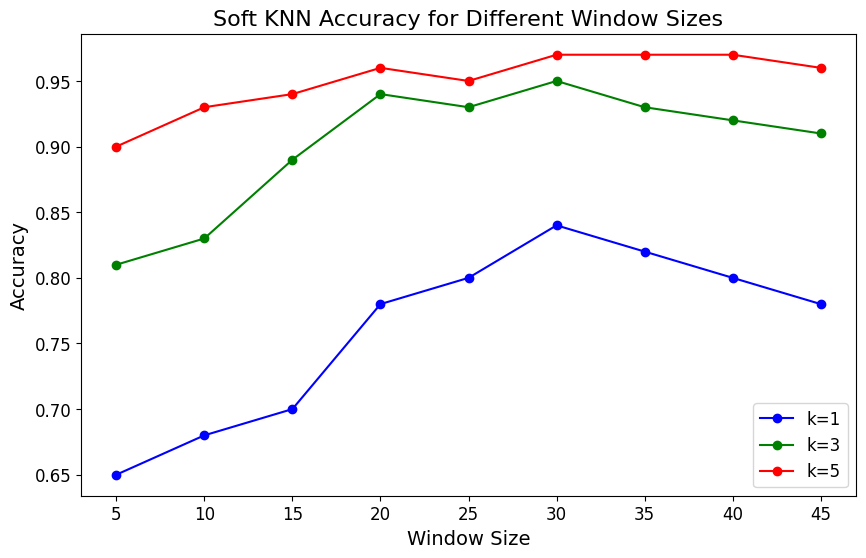}
    \end{subfigure}
    \vspace{-1mm}
    \caption{Contrastive loss (left) and soft $k$NN accuracy (right) with different \texttt{WINDOW\_SIZEs}. Note that the results for hard $k$NN (regardless of the value of $k$) are identical to that of soft $k$NN when $k=1$.}
    \label{fig:ablation}\vspace{-1mm}
\end{figure*}

\subsection{Scalability}
\label{sec:exp:scalability}
Contemporary LLMs often necessitate substantial quantities of training data and the capability to manage a multitude of data providers. Hence, it is imperative to demonstrate the scalability of the \texttt{TRACE} framework under such  settings. We assess the scalability of \texttt{TRACE} by selecting $10$, $25$, $50$, and $100$ distinct books from the \texttt{booksum} dataset, while maintaining a consistent experimental configuration. The results in Table~\ref{table: scalability} indicate a diminishing trend in accuracy with an increasing number of data providers, which is expected as the task complexity grows. However, despite this challenge, \texttt{TRACE} exhibits a relatively high level of accuracy across all settings, thus affirming its scalability.

\subsection{Interpretability}
\label{sec:exp:Interpretability}
The \texttt{TRACE} framework not only delivers accurate source attribution but also provides interpretability by offering additional insights into the attribution process. This interpretability is crucial for understanding the reasoning behind the model's decisions and gaining confidence in its responses. We illustrate the interpretability of TRACE using responses from the \texttt{t5-small-booksum} model as a demonstration.

\begin{table*}[t]
\centering
\resizebox{0.75\linewidth}{!}{
\begin{tabular}{c|ccc|ccc|c}
\toprule
\multirow{2}{*}{Inference method} & \multicolumn{3}{c|}{deletion} & \multicolumn{3}{c|}{synonym substitution} & \multirow{2}{*}{paraphrasing} \\
 & 5\% & 10\% & 15\% & 5\% & 10\% & 15\% \\
\midrule
top-1 acc. drop & ↓0.9\% & ↓1.2\% & ↓1.7\% & ↓1.5\% & ↓2.9\% & ↓3.5\% & ↓9.5\% \\
top-3 acc. drop & ↓0.7\% & ↓1.3\% & ↓1.6\% & ↓1.7\% & ↓2.3\% & ↓2.7\% & ↓4.5\% \\
top-5 acc. drop & ↓0.3\% & ↓0.7\% & ↓0.7\% & ↓1.1\% & ↓2.1\% & ↓2.3\% & ↓1.8\% \\
Nearest Centroid drop & ↓0.9\% & ↓1.2\% & ↓1.7\% & ↓1.5\% & ↓2.9\% & ↓3.5\% & ↓9.5\% \\
\bottomrule
\end{tabular}
}
\caption{Impact of various text perturbation attacks on \texttt{TRACE} attribution accuracy using \texttt{Llama-2} on the \texttt{booksum} dataset, which includes 25 distinct books. The down arrows indicate the percentage drop in accuracy due to the corresponding attack ratio.}
\label{table: robustness}
\end{table*}

Table~\ref{tab:example} shows a summary of correctly attributed single-source responses from the \texttt{t5-small-booksum} model. Each response is paired with the nearest principal sentence from the identified source. This pairing allows users to understand the specific evidence or context from the source text that influences the model's attribution decision.

Moreover, TRACE offers interpretability through the inclusion of different similarity scores. These scores provide insights into the model's confidence levels regarding the attribution outcomes. By examining the similarity scores, users can gauge the strength of the connection between the response and the identified source.

Overall, TRACE enhances interpretability by not only delivering the final attribution outcomes but also by providing supporting evidence from the source text and indicating the model's confidence levels through similarity scores. This transparency and insight into the attribution process empower users to trust and understand the model's responses, which makes \texttt{TRACE} a valuable tool for source attribution tasks.

\begin{table*}[h]
    \centering
    \resizebox{0.75\linewidth}{!}{
        \begin{tabular}{|p{0.48\linewidth}|p{0.48\linewidth}|}
            \hline
            \textbf{Response} & \textbf{Nearest Principal Sentence} \\
            \hline
            \RaggedRight Morel is in Sheffield, and he feels guilty towards Dawes, who is suffering and despairing, too. And besides, they had met in Nottingham in a way that is more or less responsible. & \RaggedRight on his knees, feeling so awkward in presence of big trouble. Mrs. Morel did not change much. She stayed in Sheffield\\
            \hline
            \RaggedRight But Emma thought at least it would turn out so. Mrs. Elton was first seen at church: but although devotion might be interrupted, curiosity could not be satisfied by a bride in. Pew, and it must be left for the visits in form which were then paid, to settle whether she was very pretty indeed, or only rather pretty at all. & \RaggedRight or any thing just to keep my boot on.'' Mr. Elton looked all happiness at this proposition; and nothing could exceed\\
            \hline
            \RaggedRight to marry Lord Warburton. Isabel enquired. ``Your uncle's not an English nobleman,'' said Mrs. Touchett in her smallest, sparest voice. The girl asked if the correspondent of the Interviewer was to take the party to London under Ralph's escort. It was just the sort of plan, she said, that Miss Stackpole would be sure to suggest, and Isabel said that she did right to refuse him then. & \RaggedRight he told Ralph he's engaged to be married.'' ``Ah, to be married!'' Isabel mildly exclaimed. ``Unless he breaks it off. He seemed \\
            \hline
        \end{tabular}
    }
    \caption{Sample responses with correct single-source attribution from \texttt{t5-small-booksum} model.}
    \label{tab:example}
\end{table*}

\subsection{Robustness}
In the context of adversarial scenarios, it is essential to evaluate the robustness of \texttt{TRACE} against attacks wherein the attacker has access to the model's response. We consider attackers who can apply distortions to the response to alter the source attribution results but have black-box access to the model itself. This section focuses on evaluating the robustness of \texttt{TRACE} under three specific types of distortions: deletion, synonym substitution, and paraphrasing. The summarized results are presented in Table\ref{table: robustness}. Appendix~\ref{app:attack} presents the attack details.

The results indicate that while deletion and synonym substitution do have some impact on attribution accuracy, the extent of this impact is minimal. However, paraphrasing proves to be a more potent attack. This is because paraphrasing essentially alters the sentences to a larger extent, which causes more influence on the source attribution results. This observation highlights the need for future research to develop effective defense mechanisms for \texttt{TRACE} against such adversarial techniques.

\subsection{Additional Experimental Results}
\label{sec:additional}
We conduct additional experiments to assess the performance of \texttt{TRACE} on alternative datasets, thereby evaluating its versatility. Table~\ref{table:additional} summarizes the results. For a consistent comparison, we employ the same LLM across these datasets. 

\begin{table*}[h]
    \centering
    \resizebox{0.75\linewidth}{!}{
        \begin{tabular}{c|c|ccc|cc|c}
            \toprule
            \multirow{2}{*}{Dataset} & \multirow{2}{*}{Data Providers} & \multicolumn{3}{c|}{Soft $k$NN} & \multicolumn{2}{c|}{Hard $k$NN} & \multirow{2}{*}{Nearest Centroid}  \\
             &  & $k=1$  & $k=3$ & $k=5$  & $k=10$ & $k=20$ &  \\
            \midrule
             \texttt{booksum} & 10 & 85.3\% & 96.8\% & 98.7\% & \multicolumn{2}{c|}{85.3\%} & 85.3\% \\ 
            \midrule
            \texttt{dbpedia\_14} & 10 & 88.2\% & 94.1\% & 97.2\% & \multicolumn{2}{c|}{88.2\%} & 88.2\% \\
            \midrule
            \hline
             \texttt{booksum} & 25 & 81.3\% & 92.3\% & 94.0\% & \multicolumn{2}{c|}{81.3\%} & 81.3\% \\ 
            \midrule
            \texttt{cc\_news} & 25 & 83.1\% & 90.8\% & 92.1\% & \multicolumn{2}{c|}{83.1\%} & 83.1\% \\ 
            \bottomrule 
        \end{tabular}
    }
    \caption{Source attribution accuracy on \texttt{dbpedia\_14} and \texttt{cc\_news} datasets using \texttt{TRACE}.}
    \label{table:additional}
    \vspace{-1mm}
\end{table*}

Our additional experiments affirm the adaptability of the \texttt{TRACE} framework across various datasets, thereby validating its applicability across various knowledge domains and settings.

\subsection{Ablation Study}
\label{sec:ablation}
The most important factor in \texttt{TRACE} is the user-defined \texttt{WINDOW\_SIZE}. If the \texttt{WINDOW\_SIZE} is too small, the principal sentences cannot capture sufficient contextual information, hence deteriorating the performance. However, an exceedingly large \texttt{WINDOW\_SIZE} will not only require more computational resources and time to train but also the meaning will be diluted by other redundant information. This presents a natural trade-off between source attribution performance and computational efficiency. Therefore, in this subsection, we will analyze this trade-off and present the results in Fig.~\ref{fig:ablation}.

It can be observed that a larger \texttt{WINDOW\_SIZE} facilitates faster model convergence. However, model loss alone is not a comprehensive indicator of the clustering quality. So, we evaluate the source attribution accuracy on the test dataset. When the \texttt{WINDOW\_SIZE} is set to $30$, our \texttt{TRACE} framework achieves its highest accuracy. We hypothesize that this is primarily because the \texttt{WINDOW\_SIZE} of $30$ is sufficient to capture essential contextual information without excessively diluting it.

\section{Related Work}

\paragraph{Source Attribution.}
Though source attribution remains relatively underexplored in the domain of natural language processing, \texttt{WASA}~\citep{wang2023wasa} stands out as a notable framework.\footnote{Note that neither the source code nor 
comprehensive details of the experimental setup have been provided in~\citep{wang2023wasa}, making a fair comparison with \texttt{WASA} infeasible.} Operating on the principle of watermarking, \texttt{WASA} embeds distinct source identifiers within the training data to ensure that responses convey pertinent data provider information. However, \texttt{WASA} necessitates extensive manipulation of training data and training the entire LLM from scratch, which is a time-consuming process given their sizes. In contrast, \texttt{TRACE} distinguishes itself by being model-agnostic, i.e., requiring no knowledge about the model. This characteristic enhances efficiency and adaptability.  

In the context of identifying information sources for quotes, \texttt{Quobert}~\citep{10.1145/3437963.3441760} is a minimally supervised framework designed for extracting and attributing quotations from extensive news corpora. Additionally, \citet{spangher-etal-2023-identifying} have developed robust models for identifying and attributing information in news articles. However, these approaches are primarily focused on specific domains such as news. In contrast, \texttt{TRACE} is designed to handle knowledge across a wide range of domains and hence provides a more generalized and versatile solution for source attribution tasks.

\paragraph{Information Retrieval.}
A related topic to our work here is information retrieval. Traditional retrieval techniques like BM25~\citep{Robertson} hinge heavily on frequency-based rules which prove to be inadequate when dealing with responses that share semantic similarities without significant lexical overlap. More contemporary methods, such as ANCE~\citep{xiong2020approximate} and Contriever~\citep{izacard2022unsupervised}, opt for generating compact, dense representations of documents rather than long, sparse ones. Thus, they tend to achieve better results. 

While information retrieval and \texttt{TRACE} both use dense representations to measure sentence similarity, they differ in objectives and applications. Information retrieval aims to rank relevant documents for a user's query. In contrast, \texttt{TRACE} focuses on identifying and attributing the original source of specific information, hence ensuring accurate credit and authenticity.

\paragraph{Membership Inference Attack.}
The concept of membership inference attack was first introduced by~\citet{7958568}. The primary objective of this attack is to ascertain whether a specific piece of information was part of the training data for a given machine learning model. Various assumptions about the available information lead to different attack models. For instance, some models assume access to hard labels~\citep{li2021membership}, the model's confidence scores~\citep{watson2022importance, mattern2023membership}, or the internal parameters of the model~\citep{255348}.~\citet{wei2024proving} have achieved membership inference by inserting watermarks into data.
Despite the variations, these attacks fundamentally seek to answer a binary question, i.e., whether the information was included in the training dataset or not.

In contrast, source attribution entails mapping the response to distinct and specific sources rather than simply determining the presence or absence of the data in the training set. Additionally, \texttt{TRACE} adheres to a black-box setting: It does not require access to internal information such as confidence scores or model parameters. Instead, \texttt{TRACE} only necessitates the response from a LLM.

\section{Conclusion}
This paper describes a novel \texttt{TRACE} framework which effectively achieves source attribution. By selecting principal sentences and projecting them into the embedding space via source-coherent contrastive learning, \texttt{TRACE} enhances the interpretability of responses generated by LLMs. This enhancement also conforms to regulations that aim to protect the privacy of users. After evaluating \texttt{TRACE} on various datasets, we have demonstrated the accuracy , scalability, and robustness of our framework.

\subsection*{Limitations}
Our experiments are subject to some limitations that can be addressed in the future work to ensure a comprehensive interpretation of results. Firstly, the balanced distribution of data across different sources may impact the final inference of \texttt{TRACE} given its reliance on the $k$NN algorithm. This uniformity in data volume may not be representative of real-world settings, which potentially limits the generalizability of our findings. Secondly, information within each source is quite distinct with no overlapping data. Future works can verify the setting where data sources contain similar information. These limitations underscore the importance of future research in addressing such challenges to enhance the robustness of \texttt{TRACE} across varied data environments.

\subsection*{Ethical Considerations}
Our \texttt{TRACE} framework introduces a method for achieving source attribution. Utilizing this framework, a malicious actor may potentially identify the sources of data providers and reveal sensitive information about them. Therefore, the application of TRACE within this context necessitates meticulous handling to mitigate privacy concerns.


\newpage
\bibliography{acl_latex}

\begin{thebibliography}{42}
\providecommand{\natexlab}[1]{#1}

\bibitem[{Arora et~al.(2019)Arora, Khandeparkar, Khodak, Plevrakis, and Saunshi}]{arora2019theoretical}
Sanjeev Arora, Hrishikesh Khandeparkar, Mikhail Khodak, Orestis Plevrakis, and Nikunj Saunshi. 2019.
\newblock A theoretical analysis of contrastive unsupervised representation learning.
\newblock {arXiv}:1902.09229.

\bibitem[{Brochier et~al.(2019)Brochier, Guille, and Velcin}]{Brochier_2019}
Robin Brochier, Adrien Guille, and Julien Velcin. 2019.
\newblock Global vectors for node representations.
\newblock In \emph{Proc. WWW}, pages 2587--2593.

\bibitem[{Cer et~al.(2018)Cer, Yang, yi~Kong, Hua, Limtiaco, John, Constant, Guajardo-Cespedes, Yuan, Tar, Sung, Strope, and Kurzweil}]{cer2018universal}
Daniel Cer, Yinfei Yang, Sheng yi~Kong, Nan Hua, Nicole Limtiaco, Rhomni~St. John, Noah Constant, Mario Guajardo-Cespedes, Steve Yuan, Chris Tar, Yun-Hsuan Sung, Brian Strope, and Ray Kurzweil. 2018.
\newblock Universal sentence encoder.
\newblock {arXiv}:1803.11175.

\bibitem[{Chen et~al.(2020)Chen, Kornblith, Norouzi, and Hinton}]{chen2020simple}
Ting Chen, Simon Kornblith, Mohammad Norouzi, and Geoffrey Hinton. 2020.
\newblock A simple framework for contrastive learning of visual representations.
\newblock {arXiv}:2002.05709.

\bibitem[{Conneau et~al.(2018)Conneau, Kiela, Schwenk, Barrault, and Bordes}]{conneau2018supervised}
Alexis Conneau, Douwe Kiela, Holger Schwenk, Loic Barrault, and Antoine Bordes. 2018.
\newblock Supervised learning of universal sentence representations from natural language inference data.
\newblock {arXiv}:1705.02364.

\bibitem[{Cui et~al.(2021)Cui, Zhong, Liu, Yu, and Jia}]{Cui_2021_ICCV}
Jiequan Cui, Zhisheng Zhong, Shu Liu, Bei Yu, and Jiaya Jia. 2021.
\newblock Parametric contrastive learning.
\newblock In \emph{Proc. ICCV}, pages 715--724.

\bibitem[{Hadsell et~al.(2006)Hadsell, Chopra, and Lecun}]{inproceedings}
Raia Hadsell, Sumit Chopra, and Yann Lecun. 2006.
\newblock Dimensionality reduction by learning an invariant mapping.
\newblock In \emph{Proc. CVPR}, pages 1735--1742.

\bibitem[{Hamborg et~al.(2017)Hamborg, Meuschke, Breitinger, and Gipp}]{Hamborg2017}
Felix Hamborg, Norman Meuschke, Corinna Breitinger, and Bela Gipp. 2017.
\newblock news-please: A generic news crawler and extractor.
\newblock In \emph{Proc. ISI}, pages 218--223.

\bibitem[{Huang et~al.(2023)Huang, Yu, Ma, Zhong, Feng, Wang, Chen, Peng, Feng, Qin, and Liu}]{huang2023survey}
Lei Huang, Weijiang Yu, Weitao Ma, Weihong Zhong, Zhangyin Feng, Haotian Wang, Qianglong Chen, Weihua Peng, Xiaocheng Feng, Bing Qin, and Ting Liu. 2023.
\newblock A survey on hallucination in large language models: Principles, taxonomy, challenges, and open questions.
\newblock {arXiv}:2311.05232.

\bibitem[{Izacard et~al.(2022)Izacard, Caron, Hosseini, Riedel, Bojanowski, Joulin, and Grave}]{izacard2022unsupervised}
Gautier Izacard, Mathilde Caron, Lucas Hosseini, Sebastian Riedel, Piotr Bojanowski, Armand Joulin, and Edouard Grave. 2022.
\newblock Unsupervised dense information retrieval with contrastive learning.
\newblock {arXiv}:2112.09118.

\bibitem[{Kry{\'s}ci{\'n}ski et~al.(2022)Kry{\'s}ci{\'n}ski, Rajani, Agarwal, Xiong, and Radev}]{kryscinski2021booksum}
Wojciech Kry{\'s}ci{\'n}ski, Nazneen Rajani, Divyansh Agarwal, Caiming Xiong, and Dragomir Radev. 2022.
\newblock {BookSum}: A collection of datasets for long-form narrative summarization.
\newblock In \emph{Proc. EMNLP Findings}, pages 6536--6558.

\bibitem[{Leino and Fredrikson(2020)}]{255348}
Klas Leino and Matt Fredrikson. 2020.
\newblock Stolen memories: Leveraging model memorization for calibrated white-box membership inference.
\newblock In \emph{Proc. SEC}, pages 1605--1622.

\bibitem[{Li and Zhang(2021)}]{li2021membership}
Zheng Li and Yang Zhang. 2021.
\newblock Membership leakage in label-only exposures.
\newblock {arXiv}:2007.15528.

\bibitem[{Liu et~al.(2024)Liu, Yao, Ton, Zhang, Guo, Cheng, Klochkov, Taufiq, and Li}]{liu2024trustworthy}
Yang Liu, Yuanshun Yao, Jean-Francois Ton, Xiaoying Zhang, Ruocheng Guo, Hao Cheng, Yegor Klochkov, Muhammad~Faaiz Taufiq, and Hang Li. 2024.
\newblock Trustworthy {LLMs}: A survey and guideline for evaluating large language models' alignment.
\newblock {arXiv}:2308.05374.

\bibitem[{Marra et~al.(2018)Marra, Gragnaniello, Verdoliva, and Poggi}]{marra2018gans}
Francesco Marra, Diego Gragnaniello, Luisa Verdoliva, and Giovanni Poggi. 2018.
\newblock Do {GANs} leave artificial fingerprints?
\newblock {arXiv}:1812.11842.

\bibitem[{Mattern et~al.(2023)Mattern, Mireshghallah, Jin, Schölkopf, Sachan, and Berg-Kirkpatrick}]{mattern2023membership}
Justus Mattern, Fatemehsadat Mireshghallah, Zhijing Jin, Bernhard Schölkopf, Mrinmaya Sachan, and Taylor Berg-Kirkpatrick. 2023.
\newblock Membership inference attacks against language models via neighbourhood comparison.
\newblock {arXiv}:2305.18462.

\bibitem[{McInnes et~al.(2020)McInnes, Healy, and Melville}]{mcinnes2020umap}
Leland McInnes, John Healy, and James Melville. 2020.
\newblock {UMAP}: Uniform manifold approximation and projection for dimension reduction.
\newblock {arXiv}:1802.03426.

\bibitem[{Meng et~al.(2021)Meng, Xiong, Bajaj, Tiwary, Bennett, Han, and Song}]{meng2021cocolm}
Yu~Meng, Chenyan Xiong, Payal Bajaj, Saurabh Tiwary, Paul Bennett, Jiawei Han, and Xia Song. 2021.
\newblock {COCO-LM}: Correcting and contrasting text sequences for language model pretraining.
\newblock {arXiv}:2102.08473.

\bibitem[{Mikolov et~al.(2013)Mikolov, Chen, Corrado, and Dean}]{mikolov2013efficient}
Tomas Mikolov, Kai Chen, Greg Corrado, and Jeffrey Dean. 2013.
\newblock Efficient estimation of word representations in vector space.
\newblock {arXiv}:1301.3781.

\bibitem[{Miller(1994)}]{miller-1994-wordnet}
George~A. Miller. 1994.
\newblock {W}ord{N}et: A lexical database for {E}nglish.
\newblock In \emph{{H}uman {L}anguage {T}echnology: Proceedings of a Workshop held at {P}lainsboro, {N}ew {J}ersey, {M}arch 8-11, 1994}.

\bibitem[{Mireshghallah et~al.(2022)Mireshghallah, Goyal, Uniyal, Berg-Kirkpatrick, and Shokri}]{mireshghallah2022quantifying}
Fatemehsadat Mireshghallah, Kartik Goyal, Archit Uniyal, Taylor Berg-Kirkpatrick, and Reza Shokri. 2022.
\newblock Quantifying privacy risks of masked language models using membership inference attacks.
\newblock {arXiv}:2203.03929.

\bibitem[{Ouyang et~al.(2022)Ouyang, Wu, Jiang, Almeida, Wainwright, Mishkin, Zhang, Agarwal, Slama, Ray, Schulman, Hilton, Kelton, Miller, Simens, Askell, Welinder, Christiano, Leike, and Lowe}]{ouyang2022training}
Long Ouyang, Jeff Wu, Xu~Jiang, Diogo Almeida, Carroll~L. Wainwright, Pamela Mishkin, Chong Zhang, Sandhini Agarwal, Katarina Slama, Alex Ray, John Schulman, Jacob Hilton, Fraser Kelton, Luke Miller, Maddie Simens, Amanda Askell, Peter Welinder, Paul Christiano, Jan Leike, and Ryan Lowe. 2022.
\newblock Training language models to follow instructions with human feedback.
\newblock {arXiv}:2203.02155.

\bibitem[{Radford et~al.(2019)Radford, Wu, Child, Luan, Amodei, and Sutskever}]{radford2019language}
Alec Radford, Jeff Wu, Rewon Child, David Luan, Dario Amodei, and Ilya Sutskever. 2019.
\newblock Language models are unsupervised multitask learners.
\newblock Technical report, OpenAI.

\bibitem[{Raffel et~al.(2020)Raffel, Shazeer, Roberts, Lee, Narang, Matena, Zhou, Li, and Liu}]{2020t5}
Colin Raffel, Noam Shazeer, Adam Roberts, Katherine Lee, Sharan Narang, Michael Matena, Yanqi Zhou, Wei Li, and Peter~J. Liu. 2020.
\newblock Exploring the limits of transfer learning with a unified text-to-text transformer.
\newblock \emph{JMLR}, 21(1):5485--5551.

\bibitem[{Reimers and Gurevych(2019)}]{reimers2019sentencebert}
Nils Reimers and Iryna Gurevych. 2019.
\newblock {Sentence-BERT}: Sentence embeddings using {Siamese} {BERT}-networks.
\newblock {arXiv}:1908.10084.

\bibitem[{Robertson et~al.(1994)Robertson, Walker, Jones, Hancock-Beaulieu, and Gatford}]{Robertson}
Stephen Robertson, Steve Walker, Susan Jones, Micheline Hancock-Beaulieu, and Mike Gatford. 1994.
\newblock Okapi at {TREC}-3.
\newblock In \emph{Proc. TREC}, pages 109--126.

\bibitem[{Schroff et~al.(2015)Schroff, Kalenichenko, and Philbin}]{Schroff_2015}
Florian Schroff, Dmitry Kalenichenko, and James Philbin. 2015.
\newblock {FaceNet}: A unified embedding for face recognition and clustering.
\newblock In \emph{Proc. CVPR}, pages 815--823.

\bibitem[{Shokri et~al.(2017)Shokri, Stronati, Song, and Shmatikov}]{7958568}
Reza Shokri, Marco Stronati, Congzheng Song, and Vitaly Shmatikov. 2017.
\newblock Membership inference attacks against machine learning models.
\newblock In \emph{Proc. IEEE S\&P}, pages 3--18.

\bibitem[{Sohn(2016)}]{NIPS2016_6b180037}
Kihyuk Sohn. 2016.
\newblock Improved deep metric learning with multi-class {N}-pair loss objective.
\newblock In \emph{Proc. NIPS}.

\bibitem[{Spangher et~al.(2023)Spangher, Peng, Ferrara, and May}]{spangher-etal-2023-identifying}
Alexander Spangher, Nanyun Peng, Emilio Ferrara, and Jonathan May. 2023.
\newblock Identifying informational sources in news articles.
\newblock In \emph{Proc. EMNLP}, pages 3626--3639.

\bibitem[{Tian et~al.(2020)Tian, Sun, Poole, Krishnan, Schmid, and Isola}]{tian2020makes}
Yonglong Tian, Chen Sun, Ben Poole, Dilip Krishnan, Cordelia Schmid, and Phillip Isola. 2020.
\newblock What makes for good views for contrastive learning?
\newblock {arXiv}:2005.10243.

\bibitem[{Touvron et~al.(2023)Touvron, Lavril, Izacard, Martinet, Lachaux, Lacroix, Rozière, Goyal, Hambro, Azhar, Rodriguez, Joulin, Grave, and Lample}]{touvron2023llama}
Hugo Touvron, Thibaut Lavril, Gautier Izacard, Xavier Martinet, Marie-Anne Lachaux, Timothée Lacroix, Baptiste Rozière, Naman Goyal, Eric Hambro, Faisal Azhar, Aurelien Rodriguez, Armand Joulin, Edouard Grave, and Guillaume Lample. 2023.
\newblock {LLaMA}: Open and efficient foundation language models.
\newblock {arXiv}:2302.13971.

\bibitem[{Vaucher et~al.(2021)Vaucher, Spitz, Catasta, and West}]{10.1145/3437963.3441760}
Timot\'{e} Vaucher, Andreas Spitz, Michele Catasta, and Robert West. 2021.
\newblock Quotebank: A corpus of quotations from a decade of news.
\newblock In \emph{Proc. WSDM}, pages 328--–336.

\bibitem[{Wang et~al.(2023{\natexlab{a}})Wang, Chen, Pei, Xie, Kang, Zhang, Xu, Xiong, Dutta, Schaeffer, Truong, Arora, Mazeika, Hendrycks, Lin, Cheng, Koyejo, Song, and Li}]{wang2023decodingtrust}
Boxin Wang, Weixin Chen, Hengzhi Pei, Chulin Xie, Mintong Kang, Chenhui Zhang, Chejian Xu, Zidi Xiong, Ritik Dutta, Rylan Schaeffer, Sang~T. Truong, Simran Arora, Mantas Mazeika, Dan Hendrycks, Zinan Lin, Yu~Cheng, Sanmi Koyejo, Dawn Song, and Bo~Li. 2023{\natexlab{a}}.
\newblock {DecodingTrust}: A comprehensive assessment of trustworthiness in {GPT} models.
\newblock In \emph{Proc. NeurIPS}.

\bibitem[{Wang et~al.(2023{\natexlab{b}})Wang, Lu, Zhao, Dai, Foo, Ng, and Low}]{wang2023wasa}
Jingtan Wang, Xinyang Lu, Zitong Zhao, Zhongxiang Dai, Chuan-Sheng Foo, See-Kiong Ng, and Bryan Kian~Hsiang Low. 2023{\natexlab{b}}.
\newblock {WASA}: Watermark-based source attribution for large language model-generated data.
\newblock {arXiv}:2310.00646.

\bibitem[{Watson et~al.(2022)Watson, Guo, Cormode, and Sablayrolles}]{watson2022importance}
Lauren Watson, Chuan Guo, Graham Cormode, and Alex Sablayrolles. 2022.
\newblock On the importance of difficulty calibration in membership inference attacks.
\newblock {arXiv}:2111.08440.

\bibitem[{Wei et~al.(2024)Wei, Wang, and Jia}]{wei2024proving}
Johnny Tian-Zheng Wei, Ryan~Yixiang Wang, and Robin Jia. 2024.
\newblock Proving membership in {LLM} pretraining data via data watermarks.
\newblock {arXiv}:2402.10892.

\bibitem[{Wu et~al.(2020)Wu, Wang, Gu, Khabsa, Sun, and Ma}]{wu2020clear}
Zhuofeng Wu, Sinong Wang, Jiatao Gu, Madian Khabsa, Fei Sun, and Hao Ma. 2020.
\newblock {CLEAR}: Contrastive learning for sentence representation.
\newblock {arXiv}:2012.15466.

\bibitem[{Xiong et~al.(2020)Xiong, Xiong, Li, Tang, Liu, Bennett, Ahmed, and Overwijk}]{xiong2020approximate}
Lee Xiong, Chenyan Xiong, Ye~Li, Kwok-Fung Tang, Jialin Liu, Paul Bennett, Junaid Ahmed, and Arnold Overwijk. 2020.
\newblock Approximate nearest neighbor negative contrastive learning for dense text retrieval.
\newblock {arXiv}:2007.00808.

\bibitem[{Yu et~al.(2022)Yu, Skripniuk, Abdelnabi, and Fritz}]{yu2022artificial}
Ning Yu, Vladislav Skripniuk, Sahar Abdelnabi, and Mario Fritz. 2022.
\newblock Artificial fingerprinting for generative models: Rooting deepfake attribution in training data.
\newblock {arXiv}:2007.08457.

\bibitem[{Zhang et~al.(2019)Zhang, Zhao, Saleh, and Liu}]{zhang2019pegasus}
Jingqing Zhang, Yao Zhao, Mohammad Saleh, and Peter~J. Liu. 2019.
\newblock Pegasus: Pre-training with extracted gap-sentences for abstractive summarization.

\bibitem[{Zhang et~al.(2015)Zhang, Zhao, and LeCun}]{NIPS2015_250cf8b5}
Xiang Zhang, Junbo Zhao, and Yann LeCun. 2015.
\newblock Character-level convolutional networks for text classification.
\newblock In \emph{Proc. NeurIPS}.

\end{thebibliography}

\newpage
\appendix

\section{Proof on the Centroid of Clusters}
\label{app:proof}
Given a cluster of embeddings $\bm{z}_1, \bm{z}_2, \ldots, \bm{z}_k$ with the same label/source, a good representative of the cluster would be the centroid $\bar{\bm{z}}$ that maximizes the sum of its cosine similarity with every normalized embedding $\bm{z}_i$ for $i = 1,\ldots,k$. Equivalently, $\bar{\bm{z}}$ minimizes the sum of its standard cosine distance with every normalized embedding: 
\[
\hspace{-1.7mm}
\begin{array}{l}
    \displaystyle\sum_{i=1}^{k} \left( 1 - \frac{\bm{z}_i \cdot \bar{\bm{z}}}{\| \bm{z}_i \| \|\bar{\bm{z}}\|} \right) 
    = k -  \sum_{i=1}^{k}\frac{\bm{z}_i \cdot \bar{\bm{z}}}{\| \bm{z}_i \| \|\bar{\bm{z}}\|} \\   
\displaystyle = k -\hspace{-0.7mm} \left( \sum_{i=1}^{k}\frac{\bm{z}_i }{\| \bm{z}_i \|} \right)\hspace{-0.7mm} \cdot\hspace{-0.7mm} \frac{\bar{\bm{z}}}{\|\bar{\bm{z}}\|}  
     \geq  k - \hspace{-0.7mm}\left\vert \left( \sum_{i=1}^{k}\frac{\bm{z}_i }{\| \bm{z}_i \|} \right) \hspace{-0.7mm}\cdot\hspace{-0.7mm} \frac{\bar{\bm{z}}}{\|\bar{\bm{z}}\|} \right\vert \\
\displaystyle\geq k - \left\| \sum_{i=1}^{k}\frac{\bm{z}_i}{\| \bm{z}_i \|} \right\|
\end{array} 
\]
by Cauchy-Schwarz inequality.
The equality holds when there exists some $\lambda \in \mathbb{R}$ such that 
\[\sum_{i=1}^{k}\frac{\bm{z}_i}{\| \bm{z}_i \|} =  \lambda \frac{\bar{\bm{z}}}{\|\bar{\bm{z}}\|}\ .
\]
In other words, $\bar{\bm{z}}$ can be obtained by adding all normalized embeddings and setting $\lambda = 1$: 
\[
\bar{\bm{z}} = \sum_{i=1}^{k}\frac{\bm{z}_i}{\| \bm{z}_i \|}\ .
\]

\section{Experimental Setup}
\label{app:train}
\paragraph{Data Preparation.}
From \texttt{booksum}, we have randomly selected subsets of $10$, $25$, $50$, and $100$ books. From \texttt{dbpedia\_14}, we chose $10$ distinct classes. Additionally, we have extracted text samples from $25$ diverse domains within the \texttt{cc\_news} dataset.

Before proceeding with the analysis, we have performed standard preprocessing steps which include converting all text to lowercase and removing punctuation to ensure uniformity and cleanliness in the data.

\paragraph{Model.}
For sentence embedding, we have opted for \texttt{SBERT}~\citep{reimers2019sentencebert}. Leveraging the pre-trained model \texttt{xlm-r-distilroberta-base-paraphrase-v1} that is readily accessible on Hugging Face, we have fine-tuned it within our \texttt{TRACE} framework. Moreover, we have augmented the model with additional feed-forward layers which serve as the projection network. The dimension for the embeddings is set as $64$.

\paragraph{Training Details.}
The hyperparameters utilized in our experimental setup are configured as follows: the learning rate is $1 \times 10^{-5}$, the batch size is $64$, the number of epochs is $150$, and the temperature in the \texttt{NT-Xent Loss} is $0.1$. Notably, all training procedures are conducted on a single NVIDIA L$40$ GPU, obviating model or data parallelism techniques. The results were obtained by averaging the outcomes of three executions, each with a different random seed.

\begin{table}[h]
\centering
\begin{scriptsize}
\begin{tabular}{lccc}
\toprule
\textbf{Statistic} & \texttt{booksum} & \texttt{dbpedia\_14} & \texttt{cc\_news} \\
\midrule
Number of Documents & 405 (books) & 560,000 & 149,954,415 \\
Languages Covered & English & English & English \\
Domains & Books & Encyclopedic & News \\
\bottomrule
\end{tabular}
\end{scriptsize}
\caption{Statistics of \texttt{booksum}, \texttt{dbpedia\_14}, and \texttt{cc\_news} datasets.}
\label{tab:dataset_stats}
\end{table}

\paragraph{Attack Details.}
\label{app:attack}
The primary attack methods utilized in this study are deletion, synonym substitution, and paraphrasing. For the deletion method, a specified portion of the response from the LLM is randomly selected and subsequently removed. In the case of synonym substitution, WordNet~\citep{miller-1994-wordnet} serves as the synonym database. Similar to the deletion method, a random portion of words is replaced with their synonyms. In deletion and synonym substitution attacks, the portions of words being modified are 5\%, 10\% and 15\%, as shown in Table~\ref{table: robustness}. For paraphrasing, we employ the fine-tuned paraphrasing model \texttt{pegasus\_paraphrase}~\citep{zhang2019pegasus}, which is available on Hugging Face.\footnote{\url{https://huggingface.co/tuner007/pegasus_paraphrase}}

\end{document}